\documentclass[10pt,twocolumn,letterpaper]{article}

\usepackage[pagenumbers]{iccv} %

\usepackage{multirow}

\newif\ifdrafting 
\draftingtrue %

\ifdrafting
    \newcommand{\maggie}[1]{{\color{violet} [Maggie: #1]}}
    \newcommand{\anurag}[1]{{\color{blue} [Anurag: #1]}}
    \newcommand{\jon}[1]{{\color{magenta} [Jon: #1]}}
    \newcommand{\rbf}[1]{{\color{red}[RBF: #1]}}
    \newcommand{\ls}[1]{{\color{cyan} [Laura: #1]}}
    \newcommand{\shreyank}[1]{{\color{green} [Shreyank: #1]}}
\else
    \newcommand{\maggie}[1]{}
    \newcommand{\anurag}[1]{}
    \newcommand{\ls}[1]{}
    \newcommand{\shreyank}[1]{}
    \newcommand{\rbf}[1]{}
    \newcommand{\jon}[1]{}
\fi
 
\newcommand{\model}{LITE }
\newcommand{\ssv}{SS-V2 }
\newcommand{\mae}{VideoMAE }
\newcommand{\sota}{state-of-the-art }
\newcommand{\oracle}{oracle }
\newcommand{\gflops}{GFLOPs }
\newcommand{\gradcam}{Grad-CAM }

\newcommand{\myparagraph}[1]{\vspace{0.2cm}
\noindent{\bf #1}}

\definecolor{cvprblue}{rgb}{0.21,0.49,0.74}
\usepackage[pagebackref,breaklinks,colorlinks,allcolors=cvprblue]{hyperref}

\def\paperID{9913} %
\def\confName{ICCV}
\def\confYear{2025}

\title{Principles of Visual Tokens for Efficient Video Understanding}

\author{Xinyue Hao\textsuperscript{1}\quad
Gen Li\textsuperscript{1}\quad
Shreyank N Gowda\textsuperscript{2}\quad
Robert B Fisher\textsuperscript{1}\quad
Jonathan Huang\textsuperscript{3}\\
Anurag Arnab\quad
Laura Sevilla-Lara\textsuperscript{1}\\ \\
\textsuperscript{1}University of Edinburgh\quad
\textsuperscript{2}University of Nottingham\quad
\textsuperscript{3}Scaled Foundations
}

\begin{document}
\twocolumn[{%
\renewcommand\twocolumn[1][]{#1}%
\maketitle
\begin{center}
    \centering
    \captionsetup{type=figure}
    \includegraphics[width=\textwidth]{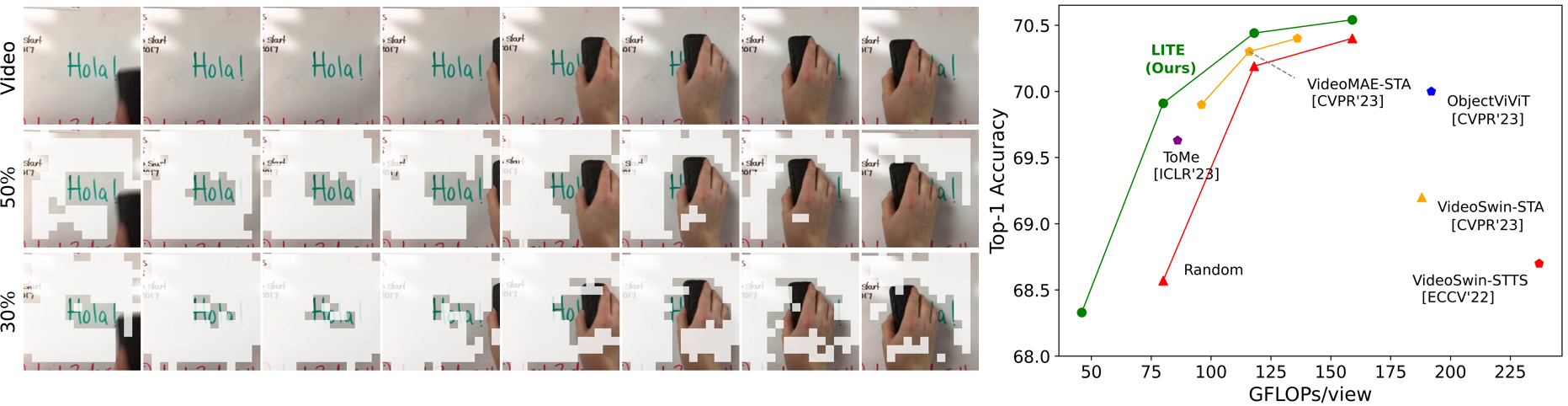}
    \captionof{figure}{Left: Given an input video (top row), the proposed \model method is able to select a subset (non-white patches) of discriminative tokens for any given computational budget. Right: \model is lightweight and efficient, achieving the optimal Pareto-front over state-of-the-art in the trade-off between computational cost (GFLOPs) and accuracy. The plot shows the results on Something-Something-V2.
    }
    \label{fig:teaser}
\end{center}%
}]

\maketitle
\begin{abstract}
Video understanding has made huge strides in recent years, relying largely on the power of transformers. As this architecture is notoriously expensive and video data is highly redundant, research into improving efficiency has become particularly relevant. Some creative solutions include token selection and merging. While most methods succeed in reducing the cost of the model and maintaining accuracy, an interesting pattern arises: most methods do not outperform the baseline of randomly discarding tokens. In this paper we take a closer look at this phenomenon and observe 5 principles of the nature of visual tokens. For example, we observe that the value of tokens follows a clear Pareto-distribution where most tokens have remarkably low value, and just a few carry most of the perceptual information. %
We build on these and further insights to propose a lightweight video model, LITE, that can select a small number of tokens effectively, outperforming state-of-the-art and existing baselines across datasets (Kinetics-400 and Something-Something-V2) in the challenging trade-off of computation (GFLOPs) vs accuracy. Experiments also show that LITE generalizes %
across datasets and even other tasks without the need for retraining. 
\end{abstract}
    
\section{Introduction}
\label{sec:intro}

\begin{figure}
    \centering
    \includegraphics[width=0.8\linewidth]{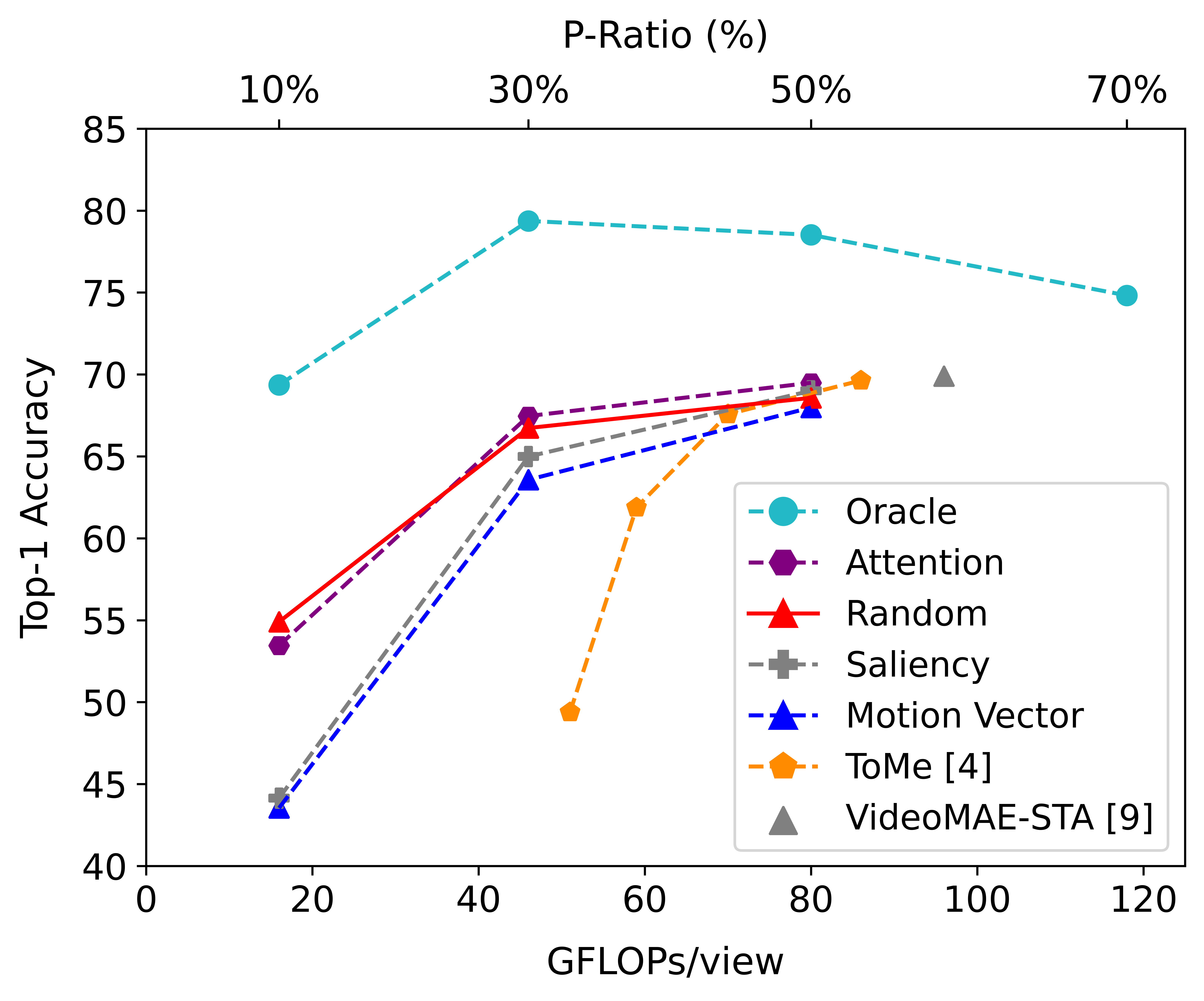}
    \caption{Random sampling of tokens is comparable to or outperforms most sophisticated models for token selection, as well as other sampling baselines. The ``P-Ratio'' represents the proportion of tokens selected relative to the total number of tokens per video. %
    }
    \label{fig:random-strong}
    \vspace{-0.4cm}
\end{figure}

Video understanding has made remarkable progress in the last few years, getting close to solving many standard benchmarks and tasks~\cite{carreira2017quo,gu2018ava,caba2015activitynet}. This progress has largely hinged on the use of the transformer architecture~\cite{vaswani2017attention}, which is both extremely powerful as well as extremely computationally expensive. Transformers originated in the language field, where few tokens are needed to represent a concept, such as an action, as tokens roughly correspond to single words. This is not the case in the visual adaptation of transformers~\cite{Arnab2021ViViTAV}, where the number of tokens to represent an action is orders of magnitude larger, leading to an exponentially larger computational cost. This cost has a wide range of negative consequences: it limits the deployability of models, as they require expensive equipment to run, it burdens video understanding research, making turnaround slower, it is financially expensive and it has a big environmental footprint~\cite{gowda2023watt}. In particular, although the inference cost is small compared to training, over the lifetime of a model, the overall cost of inference is larger than the cost of training, as it is done many more times~\cite{inferencecost}. 
At the same time, video is notoriously redundant in space (where background regions might be a large part of the scene) and time (with many frames being similar, even if they are downsampled). These two factors together pose a great opportunity for token reduction and have led a research theme with many creative ideas. One of the noticeable strands is token merging, where similar tokens are grouped together such that only a representative of the group needs to be stored in memory \cite{tome,Ding2023PruneST,choi2024vid,Feng2023EfficientVT}. Another direction includes selecting tokens based on motion~\cite{Patrick2021KeepingYE}. %
These models can reduce the computational cost in \gflops considerably, while keeping accuracy on-par with the full model. While this is promising, we observe that most token-reduction models make a trade-off between accuracy and computation that is similar or inferior to the simplest baseline of sampling tokens randomly (see Fig.~\ref{fig:random-strong}). In this paper we take a closer look at this puzzling effect, to discover underlying principles of the nature of visual tokens. We leverage these principles to design faster, more efficient video models.

\myparagraph{Using a subset of ``good" tokens can be better than using all tokens.} First, we design an oracle that estimates the value of each token for a particular task, concretely we take {\em Action Classification} as our testbed. This oracle is created such that the value of each input token corresponds to its gradient~\cite{Selvaraju2016GradCAMVE}. We refer to it as an oracle because it uses the ground truth label of the class. %
Given this oracle, we can now sample a subset of the tokens according to their gradient value, keeping those with higher values. 
We observe that the accuracy of using a subset of tokens of a video according to the oracle is actually much larger than the baseline of using all tokens. In other words, ``low-value" tokens not only do not help, but actually act as noise that hurt classification. Crucially, this gap is also shockingly large: it can be up to 9\%. This demonstrates even further the surprising value of the oracle.  
\begin{figure}
    \centering
    \includegraphics[width=0.9\linewidth]{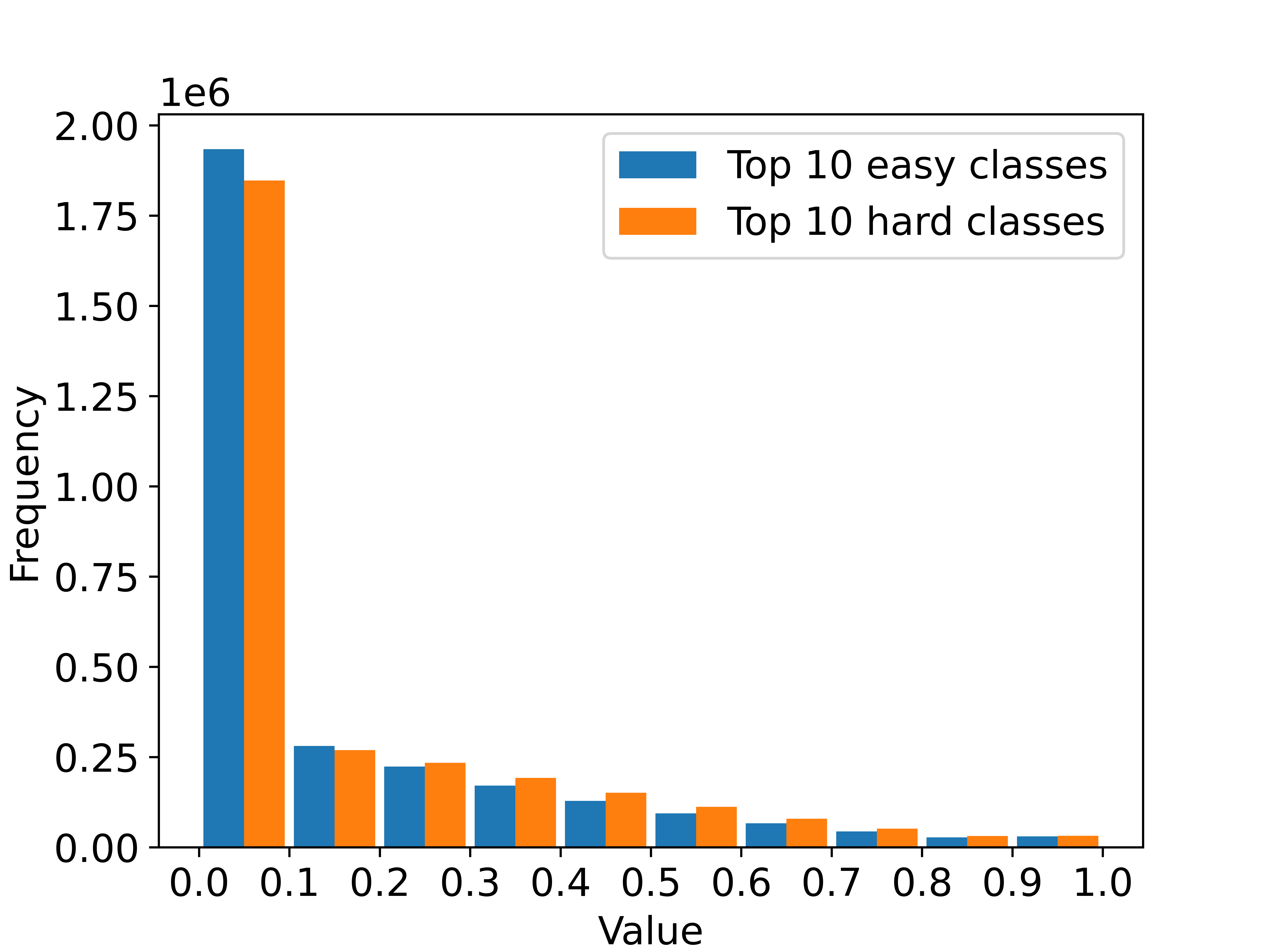}
    \caption{ Histogram of value of tokens as predicted by the oracle, showing a very clear Pareto-like distribution. Easy classes have thinner tails than hard classes.}
    \label{fig:histogram}
    \vspace{-0.5cm}
\end{figure}

\myparagraph{The value of tokens follows a Pareto distribution. } We can plot a histogram of the value of tokens estimated by our oracle and observe that it closely follows a Pareto distribution, where the vast majority of tokens carry a very small amount of information, while a select few carry most of the information necessary for classification. This helps explain the puzzling power of the random-drop baseline: when sampling randomly, we are more likely to discard from valueless pile, therefore less likely to affect the overall performance of the model. 

\myparagraph{Searching for the right tokens is like looking for a needle in a haystack.} What does the oracle's performance look like? Does it attend to cues that humans use, such as foreground objects? Furthermore, can we train a simple model to reproduce the information in the oracle? We conducted a series of experiments to understand the nature of the tokens selected by the oracle.
We assess how they compared to those selected by other methods, including: selecting the foreground, selecting the regions with more attention, etc. Surprisingly, all these strategies perform worse than selecting tokens at random. We then propose a ``selector" network, which we call \model for {\em LIghtweight Token Elector}, which takes as input a token and approximates the value of the oracle. We observe that although the selector is extremely lightweight, it is enough to outperform previous work on token selection and merging, including the strong random-selection baseline. This selector is a simple multi-layer perceptron (MLP) which operates at the token level individually, and in our experiments no 
comparison methods give better performance. 

\myparagraph{Easy tasks do not need as much computation to maintain high accuracy.} Easy classes tend to have token distributions with slightly thinner tails. This suggests that potentially there are fewer ``essential" tokens for classification. We measure the impact of adapting the computational budget on a per-video basis, such that easier videos (with high prediction confidence) use a smaller budget than harder videos (with low prediction confidence). 
\\
\noindent We make several {\bf contributions}: 
\begin{itemize}
    \item We provide a comprehensive study of the nature of value of tokens in video, distilling our findings into 5 principles that can be leveraged for future research. 
    \item We incorporate these principles into a novel method, %
    that can be used as plug-and-play for other settings.
    \item We advance the state-of-the-art in efficiency of video across standard datasets, maintaining accuracy while cutting over of GFLOPs. 
\end{itemize}

Note that this work focuses on {\em visual} tokens. This in practice means that our experiments use vision models and tasks (e.g., classification, zero-shot learning), instead of vision-language models (VLMs) and tasks (e.g., VQ\&A, captioning). There are several reasons for this choice. 
First, while VLMs have shown excellent progress, much of their impressive nature is due to progress in language technology (e.g. LLMs and language embeddings) \cite{xiao2024can}. We find it urgent and essential to advance knowledge in the understanding of visual tokens, and how that will affect the development of future efficient and effective visual representations.
Second, the distribution and properties of visual tokens from Vision Transformer (ViT) and vision-language tokens from VLMs are not necessarily similar. %
In particular, vision-language tokens in VLMs are often learned in a contrastive manner, to fit to the representation and distribution of language tokens which are different in nature~\cite{chen2024image}. %
Therefore we would not expect conclusions and properties of both families of tokens to be necessarily interconnected. We do experiment extensively with the zero-shot setting of datasets and tasks (which is one of the appealing properties of VLMs) and observe that the proposed model~\model successfully applies to novel datasets and tasks without the need for retraining.

\section{Related Work}
\label{sec:related_work}
\vspace{-0.2cm}
\myparagraph{Video Transformers.} Transformer-based architectures \cite{vaswani2017attention} have generally delivered outstanding improvements in various image-related tasks \cite{dosovitskiy2020image, touvron2021training, liu2021swin}. Beyond images, they have also spurred significant research in video understanding \cite{Arnab2021ViViTAV,bertasius2021space, Bulat2021SpacetimeMA,Fan2021MultiscaleVT,liu2021swin,Neimark2021VideoTN,Patrick2021KeepingYE,sfa-vivit,wasim2023video}, yielding promising outcomes. However, the computational cost of Transformers increases quadratically with the number of tokens, which presents challenges for handling longer videos. 

\myparagraph{Efficient Video Recognition.}
Video Understanding has struggled with efficiency for years, spurring a big body of work on reducing input data~\cite{Wu2017CompressedVA,Zolfaghari2018ECOEC,Feichtenhofer2018SlowFastNF,Feichtenhofer2020X3DEA,Kondratyuk2021MoViNetsMV}, leveraging motion~\cite{Patrick2021KeepingYE}, %
or using a memory cache~\cite{Wu2022MeMViTMM} to enable long-video analysis. Some recent work leverage advances in image models to adapt them to videos using parameter-efficient fine-tuning~\cite{aim,st-adapter,actionclip,x-clip}.

\myparagraph{Transformer Token Reduction.} 
 From all efficiency strategies, the most relevant to this work is reduction of tokens. %
 Some methods drop tokens by predicting scores through end-to-end training \cite{rao2021dynamicvit, Yin2021AViTAT, Wang2021EfficientVT, liu2024revisiting}. For instance, DynamicViT~\cite{rao2021dynamicvit} scores each token using a lightweight network and selects top tokens with Gumbel-softmax. A-ViT~\cite{Yin2021AViTAT} computes halting scores from the token embedding’s first dimension to decide on pruning, while STTS~\cite{Wang2021EfficientVT} uses a differentiable Top-$K$ operator to predict scores and select top areas for reduced computation.

Additionally, some methods extract essential information from tokens to reduce their number~\cite{Park2022KcenteredPS, ryoo2021tokenlearner, Koner2024LookupViTCV}. Strategies like $K$-centered search~\cite{Park2022KcenteredPS} are used for this purpose. TokenLearner~\cite{ryoo2021tokenlearner} employs MLPs to learn a fixed number of new tokens to reduce the computational load. LookupVit~\cite{Koner2024LookupViTCV} uses compressed tokens for high computational processing, and other tokens for cheaper layers.

Another mainstream approach is merging redundant or similar tokens~\cite{tome,Ding2023PruneST,choi2024vid,Feng2023EfficientVT}. The seminal ToMe~\cite{tome} method merges similar tokens using bipartite matching algorithms. Building on ToMe, STA~\cite{Ding2023PruneST} developed a video-specific token dropping method where the first frame uses ToMe, and each subsequent frame is compared with the remaining tokens from the previous frame, retaining similar ones. Additionally, the vid-TLDR~\cite{choi2024vid} work incorporates a saliency mask based on ToMe, enhancing the contribution of tokens in salient areas in subsequent learning. Moreover, the recent STTM~\cite{Feng2023EfficientVT} method extracts object cues, compresses token information within the same object, and add aggregated position information to compensate for the loss of motion cues. A special case of token reduction is frame selection, which has led to a series of interesting work including SMART~\cite{gowda2021smart}, Sevila~\cite{yu2024self}, Vila~\cite{wang2024vila}, and ATP~\cite{buch2022revisiting}. 

Unlike previous data-centric approaches, LITE uses an oracle-driven, model-focused method to retain high-value tokens via direct gradient scoring, adapting compute based on video complexity to improve efficiency and accuracy over existing methods.

\vspace{-0.1cm}
\section{Principles of Visual Tokens in Video}
\label{sec:analysis}
\vspace{-0.1cm}

In this section we take a closer look at the nature of visual tokens in video. The goal is to use the underlying principles to design an efficient model that can both reduce the computational complexity of models at test time as well as leverage the overwhelming redundancy of tokens in video. %
We introduce the experimental setup, including architectures, datasets, and implementation details used throughout the paper, followed by key principles and supporting experimental evidence.

\myparagraph{Datasets.} We use three standard action recognition datasets and one spatial-temporal action detection dataset:
Kinetics-400 (K400) \cite{carreira2017quo}, sourced from YouTube, is a large-scale human action video dataset with 400 classes. Each class has a minimum of 400 action clips.
Something-Something V2 (SS-V2) \cite{Goyal2017TheS} focuses on fine-grained understanding of human hand gestures. Comprising 220,847 labeled video clips, it captures 174 individuals performing predefined actions with everyday items. This dataset poses a greater challenge for temporal modeling \cite{SevillaLara2019OnlyTC}.
UCF101 \cite{soomro2012ucf101}  is a dataset collected from YouTube, consisting of 13,320 video clips categorized into 101 different action classes.
AVA Actions (AVA) \cite{gu2018ava} offers annotations for 80 atomic visual actions across 430 movie clips, each 15 minutes long. It totally provides 1.62M accurate spatial-temporal action labels.

\myparagraph{Implementation Details.} 
We use VideoMAE \cite{Tong2022VideoMAEMA} as backbone as it is very widely used. 
Frames in each video are uniformly sampled and clips of size \(16 \times 224 \times 224\) pixels are used as network inputs with a tube size of \(2 \times 16 \times 16\). During inference, we use two temporal clips. For each temporal clip, we take three spatial crops (top-left, center, bottom-right) of size \(224 \times 224\) pixels. The final prediction is obtained by averaging the scores from all views (temporal clips \(\times\) spatial crops). All experiments were carried out using 2 NVIDIA RTX 3090 24GB GPUs. 

\subsection*{Principle 1: Random is better than most}

First, we compare
the current state-of-the-art  to random  token selection, using the \ssv dataset and the \mae architecture.
Figure~\ref{fig:random-strong} shows the trade-off between computing resources (GFLOPs) and accuracy. %
In this paper, we define P-Ratio as the percentage of tokens selected compared to the total number of tokens per video.
Although recent work on token selection leverages many different cues to select discriminative tokens, surprisingly, the random baseline is comparable or superior to others like ToMe~\cite{tome} or video-specific models \cite{Ding2023PruneST} across all computational budgets. 

\subsection*{Principle 2: Good tokens do not coincide with visual cues}
It would seem that if selecting random tokens produces strong results, inserting some amount of bias should improve results.
This would also show us what type of visual content correlates more with the important parts of the scene: is it foreground? Is it moving regions? 
We experiment with several possible baselines and show results shown in Fig.~\ref{fig:random-strong}: 
\begin{itemize}
    \item {\bf Attention:} We consider using the magnitude of the attention of each token at test time. Given that these are the cues that the model is actually ``attending to" for classification, it seems like a strong signal to find good tokens. Still, it is remarkable that this is very much on-par with the random baseline. 
    \item {\bf Motion Vectors:} Motion tends to carry information about what are the discriminative regions of an action. Typically, regions that move tend to correspond to the foreground and therefore to the region where the action is taking place. In this baseline, we simply use the motion vectors (MVs) of the video, which come for free with the compressed video, and sample those that have the highest magnitude of motion. This baseline is also worse, especially for the lower computational budgets. 
    \item {\bf Saliency:}  We use a saliency detector CPFE \cite{zhao2019pyramid} to compute the saliency at each frame and sample the tokens with highest saliency. 
    It's not better than random. 
\end{itemize}

\subsection*{Principle 3: Low-value tokens can hurt recognition}

Can we characterize good tokens and bad tokens? This would help us selecting them to reduce computational complexity. However, as just shown,
reasonable inductive biases based on visual cues, such as motion, attention and saliency %
perform worse than the random selection of tokens. 
Here we try a different approach: what if we can use the gradient of each token in the video as a potential cue? This should show what parts of the video correlate more strongly with each specific class. %

\myparagraph{Designing an oracle to predict the value of tokens.} 
To identify which tokens play a decisive role in the classification task and to assign a score to each token based on their importance, we design an oracle to predict the value of a given token, using \gradcam~\cite{Selvaraju2016GradCAMVE}. As mentioned before, we call this an oracle because it has access to privileged information---the true class label. However, this is indeed a tractable approximation of a ``real oracle", which would be
to find the $k$-subset of tokens that minimizes the loss relative to the true label.

We first calculate the gradient of the score for the target class $c$, with respect to the feature map activations of the MLP in the last blocks:

\begin{equation}
\omega_d^c=\frac{1}{N} \sum_t \sum_h \sum_w \ \frac{\partial y^c}{\partial A_{t h w}^d}
\end{equation}

\noindent
where $y^c$ is the score before applying softmax for target class $c$, $A_{t h w}^d$ is the feature map activations at position $\{t,h,w\}$ with the feature $d$. After obtaining the gradient score, we perform average pooling across all tokens to derive an importance weight $\omega_d^c$ for each feature of the target class $c$. 
To approximate the ground truth as closely as possible and create a more accurate oracle, the target class selected here is the true label class.

After obtaining the importance weights for each feature, we calculate the importance score of each token for the target class using a linear combination. We incorporate a ReLU function when deriving the final token score to ensure that only tokens with a positive impact on the class are considered, making the activation maps clearer and concentrated:
\begin{equation}
S^c=ReLU\left(\sum_d \omega_d^c A^d\right)
\end{equation}
\noindent
where $A^d$ is the feature map activations with the feature $d$, $\omega_d^c$ is the importance weight of feature $d$ at target class $c$, and $S^c\in \mathbb{R}^{t\times h \times w}$ is the final token score for the target class $c$. Finally, we apply min-max normalization to scale all patch scores to a range between 0 and 1.

\begin{figure}
    \centering
    \includegraphics[width=0.9\columnwidth]{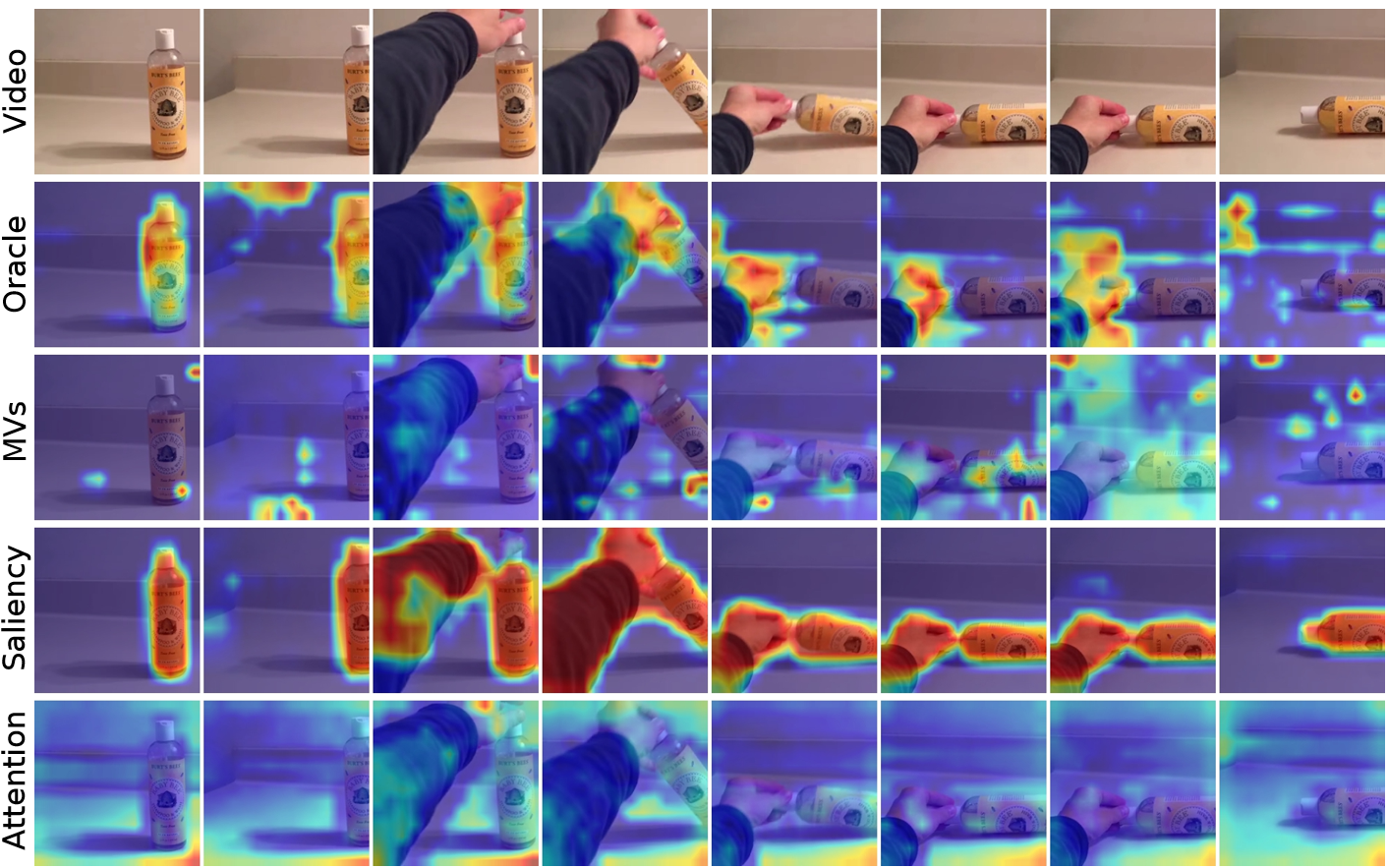}
    \caption{Sample heatmaps for different visual cues. From top to bottom: input video frames, oracle, magnitude of motion vectors, saliency, and attention. }
    \label{fig:sample_masks}
    \vspace{-0.4cm}
\end{figure}

\myparagraph{Testing the oracle.}
Is $S^c$ a good oracle? 
We measure its accuracy by selecting the top-$K$ tokens with highest value for several different computation budgets. We use \ssv as dataset and \mae as the backbone. %
Results are shown in Fig.~\ref{fig:random-strong}. We make two observations: first the oracle is much better than all other baselines, as well as the state-of-the-art. 
This is not necessarily surprising %
but it is reassuring that it is indeed a good way to measure the value of a token for recognition. We also observe that unlike other methods and baselines, the accuracy of the oracle does not strictly increase as we include more tokens. Instead, there is a sweet-spot, where high-value tokens have been chosen, but low-value, potentially noisy and confusing tokens have not been chosen. This is also an interesting insight, which gives us hope that this type of oracle can be instrumental for the design of efficient models. Figure~\ref{fig:sample_masks} shows some sample masks of what the values of the oracle look like. %

\subsection*{Principle 4: The value of tokens follows a Pareto distribution}

Figure~\ref{fig:histogram} 
shows a histogram of the oracle's token values
for a large set of videos.
This has a very Pareto-like distribution where most tokens carry little information and a few vital ones carry most of it. This helps explain the puzzling effectiveness of the random sampling: as long as some of the tail tokens are included in the random sample, the video can be classified correctly. 
Observing the shape of the accuracy curve for the random selector in Fig.~\ref{fig:random-strong}, it is far from linear. We can see that there is a cliff when the percentage of sampled tokens is smaller than 30\%. A plausible explanation is that at that point, the probability of sampling tail tokens becomes too small, and therefore it is significantly less likely that the video is classified correctly.

\subsection*{Principle 5: Easy videos require less compute}

\begin{figure}
    \centering
    \includegraphics[width=0.8\columnwidth]{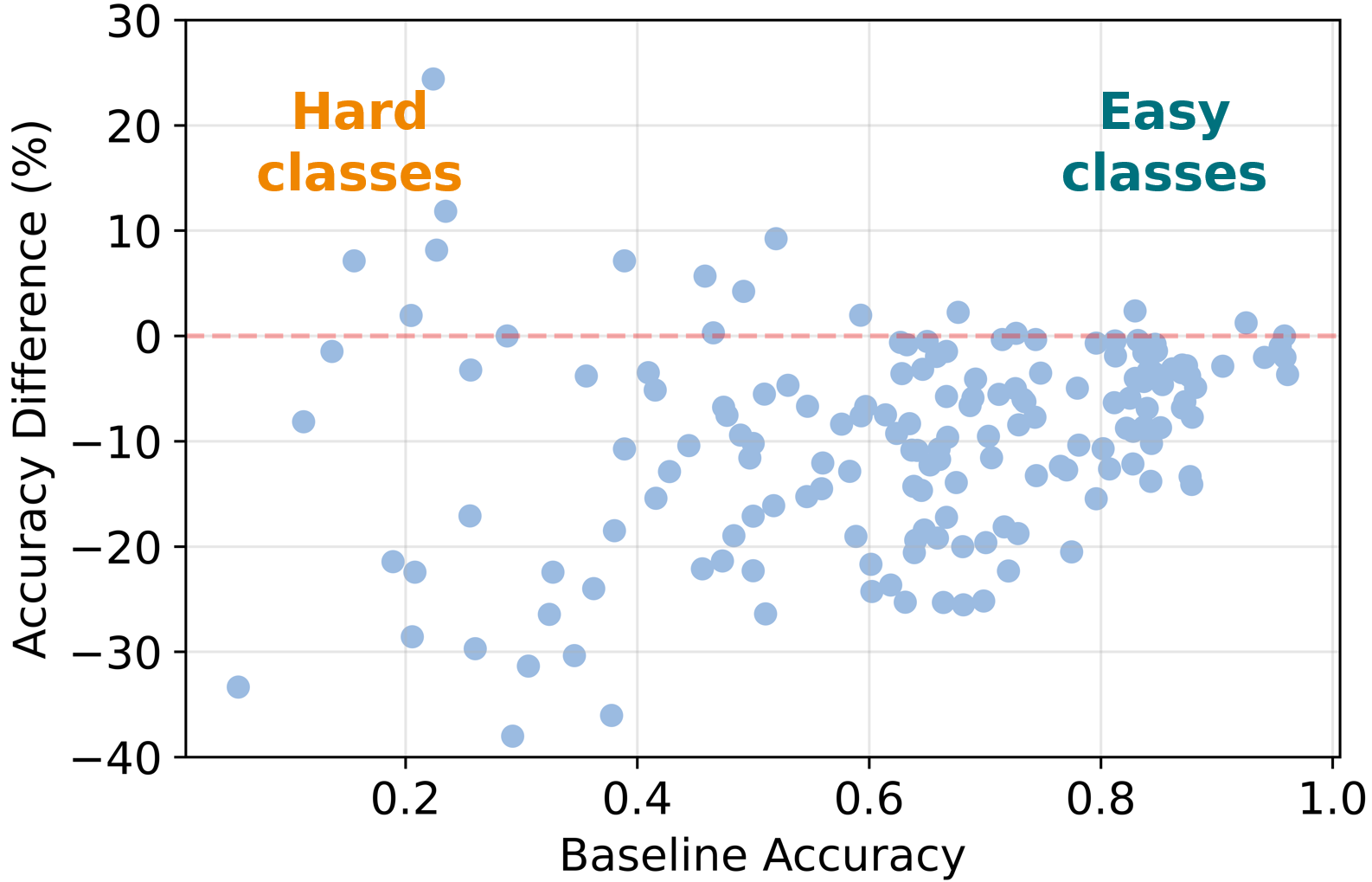}
    \caption{Impact of dropping 70\% of tokens in easy classes vs hard classes. 
    The X-axis is the accuracy when using all tokens. The Y-axis is the decay of the accuracy when dropping 70\% of tokens, as a percentage. Easier classes suffer less when the computational budget is reduced. %
    }
    \label{fig:scatter_acc}
\end{figure}

Figure~\ref{fig:histogram} shows an additional insight: easier classes have thinner tails than harder classes. Even if the difference seems small, it is statistically significant, as the number of tokens analyzed is very large. 
This can potentially mean that there are greater numbers of important or essential tokens in the hard videos than in the easy ones, and therefore they might benefit from increasing the number of sampled tokens. To explore this hypothesis we  plot, for each class, how the accuracy decreases with random sampling as a function of the original accuracy. This is seen in Fig.~\ref{fig:scatter_acc}. We observe that for a given sampling ratio (50\%), the accuracy of easier classes ({\it e.g.} those with higher baseline accuracy) is less affected than the accuracy of harder classes.  

\begin{figure*}[t]
  \centering
   \includegraphics[width=0.8\linewidth]{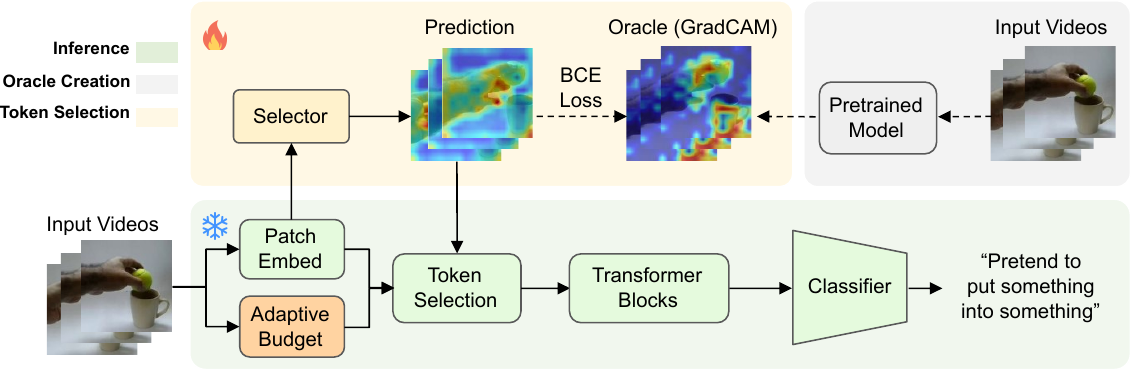}

   \caption{{\bf Overview of LITE}. At training time,  video gradients are used to build the oracle, which the selector learns to replicate. At test time, the predicted value of the selector is used to keep the top most useful tokens, reducing the overall burden of the input to the classifier. Note that the dotted line connections are only performed during training.
   }
   \label{fig:overall_structure}
   \vspace{-0.4cm}
\end{figure*}

\section{\model}
\label{sec:approach}

The findings of Sec.~\ref{sec:analysis} point to two fundamental ideas that we can leverage to design a video model, LITE, that is lightweight, efficient and adaptive and outperforms the strong random baseline as well as the previous state-of-the-art. 
These two fundamental ideas are 1) that the oracle is a reliable source of information about the value of a token and 2) that easy videos need less compute than hard videos. In this section we first review the standard Video Transformer model, which we use as backbone, and then describe how we integrate our two core ideas into \model and show an overview of the model in Fig.~\ref{fig:overall_structure}. We then discuss other possible variants that we have experimented with.

\subsection{Review of Video Transformer}
\label{sec:ReviewViT}

The Video Transformer takes as input a video clip $\mathcal{V} \in \mathbb{R}^{T \times H \times W \times 3}$, and splits them into $N$ non-overlapping patches $\mathbf{v}_n^t \in \mathbb{R}^{p_t \times p_h \times p_w}$, where $n \in 1...n_s$, $n_s =  \lvert \frac{H}{p_h} \rvert \times \lvert \frac{W}{p_w} \rvert$, $t \in 1...n_t$, $n_t = \lvert \frac{T}{p_t} \rvert$ and $N = n_s \times n_t$. These patches are converted into embeddings $\mathbf{p}_n^t \in \mathbb{R}^D$ by means of a learnable linear projection and a separate addition of spatial and temporal positional embedding, as in: 
\begin{equation}
\mathbf{p}_n^t=M \mathbf{v}_n^t +\mathbf{e}_n+\mathbf{e}^t,   
\end{equation}

where $M$ is a learnable linear matrix and $e_n$ and $e^t$ are the separate 
spatial and temporal learnable positional embeddings. These embeddings are then input to transformer blocks that perform self-attention. After that, these tokens are passed through a feedforward network, a 3-layer MLP that enables information exchange across channels.

\subsection{Learning to Select}
\label{subsec:selector}

The goal at this stage is to consider the distribution of tokens $\mathbf{p}$ in the previous section and design a lightweight model that can predict the value of the oracle for each token. Based on this prediction, we can select the tokens with highest value to input to the transformer blocks. %
A necessary condition is that the model has to be lightweight, as it has to be significantly faster than using all tokens. 

Assume a sequence of patch embeddings $\mathbf{p}$. %
To maintain simplicity and ensure the model remains lightweight, we used a three-layer fully connected multi-layer perceptron (MLP) network to predict token scores. Given that the scores $s$ from the \oracle are in the range $[0,1]$, we incorporate a sigmoid function for better estimation:
\begin{equation}
\mathbf{\hat{s}}=\operatorname{Sigmoid}\left(\mathrm{MLP}\left(\mathbf{p} ; \boldsymbol{\theta}\right)\right),
\end{equation}

\noindent
where $\mathbf{p}$ is the patch embedding, $\boldsymbol{\theta}$ is the network weights, and $\mathbf{\hat{s}}$ is the sequence of all estimated tokens score.

We train our selector in conjunction with the frozen patch embedding module from the backbone model, using binary cross-entropy loss between the selector's and oracle's token values. It is shown in the yellow region of Fig.~\ref{fig:overall_structure}.

During inference, $\mathbf{\hat{s}}$ is computed, which contains the estimated score $\mathbf{\hat{s}}(n,t)$ for each embedding $\mathbf{p}_n^t$. For a given budget $b$ given as a percentage of the total number of tokens, we select the tokens with the highest scores. 

This selector is inserted between the patch embedding module and the first transformer block, using all the weights of the original backbone model except those of our selector. This is depicted in the green region of Fig.~\ref{fig:overall_structure}. 

\subsection{Adaptive Computational Budget}

Principle 5 shows that not all videos or tasks need the same amount of computation, and easy ones need a smaller number of tokens. This is similar to the case of image recognition in human perception, where we perform certain tasks 
faster than harder tasks \cite{humanadaptive}.
Here we explore the use of this principle as follows: we gauge the difficulty of each video in a very lightweight manner and use this information to compute an ``adaptive budget", which will adapt the computational budget and save computation on ``easy" videos. %
This process is depicted as the orange box in Fig.~\ref{fig:overall_structure} and it is referred to in experiments as LITE++, as it focuses on further reducing computation at a minimal cost. 

Concretely, we gauge difficulty using confidence of the prediction as a proxy %
for whether a particular video will be correctly classified. 
To estimate this confidence we use a lightweight model (MoviNet~\cite{Kondratyuk2021MoViNetsMV}), which is considerably less accurate but extremely fast. Given the estimated confidence $\hat{c}$, the budget will be selected as follows: for very low confidence ($\hat{c} < \tau_1$) it is accepted that this will be a very hard case, and will not invest extra tokens; for easy cases ($\hat{c} > \tau_2$) we use a small budget, as it is likely that they will be easy to get right; for everything in between we use the baseline budget. 
In our experiments we use the following hyperparameters: $\tau_1 = 0.1$ and $\tau_2 = 0.5$.
For easy cases, we select top 30\% of tokens when the P-Ratio is 0.5, 0.7, or 0.9, and top 20\% of tokens when the P-Ratio is 0.3.

\subsection{Model Variants}
\label{subsec:variants}

Given the simplicity of the MLP it would appear that performance could benefit from a more sophisticated model. We experiment extensively with three families of variants: 1) improving the oracle in different ways, 2) including global information that can allow the selector to choose tokens based on their {\em joint} value, and 3) adding complexity and capacity to the selector, for example using transformer blocks or adding layers. We report results for all these variants in the Supplementary Material and observe that none of them has an impact on the accuracy of the downstream task. This points to an interesting conclusion: the \oracle is extremely hard to predict, and the MLP achieves a balance between accuracy and avoiding overfitting.

\vspace{-0.1cm}
\section{Experiments}
\label{sec:experiments}

The experimental validation on \model first compares current \sota models on the classification task (Sec.~\ref{sec:main_res}), using the mainstream datasets Kinetics-400 and SS-V2 and observe that \model achieves the pareto-front of the trade-off between computation and accuracy. We then evaluate the generalization capabilities of \model on the zero-shot setting (Sec.~\ref{sec:zero-shot}) by testing a fixed selector (trained on K400) on other datasets (UCF101~\cite{soomro2012ucf101} and SS-V2) without training or fine-tuning on them. We observe that accuracy is barely affected, showing that important tokens are consistent across domains. We take one step further and evaluate \model on a new task (detection) and new dataset (AVA~\cite{gu2018ava}) also without training. We again observe that mAP is barely affected while computation is reduced. We show extensive ablation experiments to understand the behavior of \model (Sec.~\ref{sec:ablation}). %
Overall we observe that \model outperforms previous state-of-the-art, generalizes well to new datasets and tasks and that despite its simplicity is robust and effective.

\subsection{Comparison to State-of-the-art}
\label{sec:main_res}

Tables~\ref{tab:sota_ssv2} (also shown in Fig.~\ref{fig:teaser}) and~\ref{tab:sota_k400} show the results of the balance between \gflops and Top-1 accuracy for \model and \sota on \ssv and Kinetics-400 respectively. We make several observations. %

First, \model offers a better trade-off between computational cost and accuracy than prior methods. In other words: for a fixed accuracy, \model requires less compute, and for a fixed compute, it achieves higher accuracy. For example, on the \ssv dataset, \model uses only 40\% of ObjectViViT~\cite{Zhou2023HowCO}'s \gflops with similar accuracy. On Kinetics-400, \model achieves comparable accuracy with 12\% of the \gflops used by LookupViViT~\cite{Koner2024LookupViTCV}. \model also outperforms token-merging methods: compared to ToMe~\cite{tome}, it achieves similar \gflops reduction with an accuracy gain of ~0.3\% on \ssv and 0.5\% on Kinetics-400. Likewise, \model slightly outperforms STA \cite{Ding2023PruneST} at equivalent GFLOPs.

Second, \model significantly reduces computation while maintaining accuracy, cutting \gflops by over 50\% with only a 0.6\% accuracy loss on SS-V2---an excellent trade-off. Third, this pattern is consistent; on Kinetics-400, \model saves over 50\% \gflops with a 0.9\% drop in accuracy.

\begin{table}
  \centering
  \resizebox{0.48\textwidth}{!}{%
  \begin{tabular}{lccccc}
    \toprule
    Category & Model & \gflops $\times$ \text{views} & Top-1 & Top-5 \\
    \midrule
    \multirow{4}{*}{\centering Backbones} & TimeSformer-L \cite{Bertasius2021IsSA} & 5549x1x3 & 62.40 & -- \\
    & Motionformer-L \cite{Patrick2021KeepingYE} & 1185x1x3 & 68.10 & -- \\
    & VideoSwin-B \cite{Liu2021VideoST} & 321x1x3 & 69.60 & -- \\
    & VideoMAE \cite{Tong2022VideoMAEMA} & 181x2x3 & 70.50 & 92.35 \\
    \midrule
    \multirow{7}{*}{\centering Efficient} & VideoSwin-STTS \cite{Wang2021EfficientVT} & 237x1x3 & 68.70 & -- \\
    & VIT-B-STTM \cite{Feng2023EfficientVT} & 345x1x1 & 56.50 & 83.90 \\
    & ObjectViViT \cite{Zhou2023HowCO} & 192x2x3 & 70.00 & -- \\
    & $\text{ToMe}$ \cite{tome} & 86x2x3 & 69.63 & -- \\
    & $\text{VideoMAE-STA}$ \cite{Ding2023PruneST} & 116x2x3 & 70.30 & -- \\
    & $\text{VideoSwin-STA}$ \cite{Ding2023PruneST} & 188x2x3 & 69.20 & -- \\
    & LookupViViT \cite{Koner2024LookupViTCV} & 376x4x3 & 59.60 & -- \\
    \midrule
    \multirow{4}{*}{\centering Proposed} & $\text{VideoMAE-LITE}_{90}$  & 159x2x3 & 70.54 & 92.38 \\
    & $\text{VideoMAE-LITE}_{70}$  & 118x2x3 & 70.44 & 92.30 \\
    & $\text{VideoMAE-LITE}_{50}$  & 80x2x3 & 69.91 & 91.99 \\
    & $\text{VideoMAE-LITE}_{30}$  & 46x2x3 & 68.33 & 90.87 \\
    \bottomrule
  \end{tabular}%
  }
  \caption{ Comparison to \sota on SS-V2. The subscript on LITE refers to the percentage of tokens retained, equivalent to ($\text{P-Ratio} \times 100$). The ``P-Ratio'' represents the proportion of tokens selected relative to the total number of tokens per video. }
  \label{tab:sota_ssv2}
  \vspace{-0.3cm}
\end{table}

\begin{table}
  \resizebox{0.48\textwidth}{!}{%
  \begin{tabular}{lccccc}
    \toprule
    Category & Model  & \gflops $\times$ \text{views} & Top-1 & Top-5 \\
    \midrule
    \multirow{4}{*}{Backbones} & TimeSformer-L \cite{Bertasius2021IsSA} & 8353x1x3 & 80.70 & -- \\
    & Motionformer-L \cite{Patrick2021KeepingYE} & 1185x10x3 & 80.20 & -- \\
    & ViViT \cite{Arnab2021ViViTAV} & 3981x4x3 & 84.90 & -- \\
    & VideoMAE \cite{Tong2022VideoMAEMA}  & 181x5x3 & 81.25 & 94.99  \\
    \midrule
    \multirow{5}{*}{Efficient} & VideoSwin-STTS \cite{Wang2021EfficientVT} & 253x4x3 & 81.90 & -- \\
    & VIT-B-STTM \cite{Feng2023EfficientVT}  & 345x1x1 & 73.30 & 90.50 \\
    & $\text{ToMe}$ \cite{tome} & 86x5x3 & 79.95 & -- \\
    & $\text{VideoMAE-STA}$ \cite{Ding2023PruneST} & 116x5x3 & 81.10 & -- \\
    & LookupViViT \cite{Koner2024LookupViTCV} & 376x4x3 & 78.30 & -- \\
    \midrule
    \multirow{4}{*}{Proposed} & $\text{VideoMAE-LITE}_{90}$  & 159x5x3 & 81.33 & 94.95 \\
    & $\text{VideoMAE-LITE}_{70}$ & 118x5x3 & 81.14 & 94.89 \\
    & $\text{VideoMAE-LITE}_{50}$  & 80x5x3 & 80.36 & 94.51 \\
    & $\text{VideoMAE-LITE}_{30}$ & 46x5x3 & 78.39 & 93.28 \\    
    \bottomrule
  \end{tabular}%
  }
  \caption{Comparison to \sota on Kinetics-400. The subscript in LITE refers to the percentage of tokens retained. }%
  \label{tab:sota_k400}
  \vspace{-0.5cm}
\end{table}

\subsection{Zero-shot across Datasets and Tasks}
\label{sec:zero-shot}

To evaluate the generalization capability of \model, we conduct two sets of experiments. First, we test LITE trained on Kinetics-400 on  \ssv and UCF101 to assess its ability to generalize across different datasets within the same task. 
The results of Table~\ref{tab:zeroshot} demonstrate LITE's strong zero-shot capability, even to \ssv which is visually very different from Kinetics-400. Compared to our results in Table~\ref{tab:sota_ssv2}, we observe that \model exhibits strong cross-domain generalization, achieving performance close to that of a selector trained from scratch. Second, we examine the performance of the same \model model trained on classification on an even more challenging setting: an unseen task (action detection) and unseen dataset (AVA) to evaluate its cross-task transferability. 
Table~\ref{tab:zeroshot} highlights LITE's strong cross-task generalization ability, reducing GFLOPs by 35\% while maintaining a minimal 0.7\% drop in mAP.

\begin{table}
  \centering
  \resizebox{0.48\textwidth}{!}{%
  \begin{tabular}{ccccccc}
    \toprule
     Test  & Train & Model & \gflops & Top-1 / mAP \\
    \midrule
     \multirow{3}{*}{\centering \shortstack{Classification \\ UCF101}} &\multirow{3}{*}{\centering \shortstack{Classification \\ K400}} & VideoMAE \cite{Tong2022VideoMAEMA} & 181 & 91.25 \\
     & & $\text{VideoMAE-LITE}_{90}$  & 159 & 91.33 \\
     & & $\text{VideoMAE-LITE}_{70}$  & 118 & 90.85 \\
    \cmidrule(lr){1-5}
     \multirow{3}{*}{\centering \shortstack{Classification \\ SS-V2}} &\multirow{3}{*}{\centering \shortstack{Classification \\ K400}} & VideoMAE \cite{Tong2022VideoMAEMA} & 181 & 70.50  \\
     & & $\text{VideoMAE-LITE}_{90}$  & 159 & 70.28 \\
     & & $\text{VideoMAE-LITE}_{70}$  & 118 & 69.61 \\
    \midrule
    \multirow{3}{*}{\centering \shortstack{Detection \\ AVA}} &\multirow{3}{*}{\centering \shortstack{Classification \\  K400}} & VideoMAE \cite{Tong2022VideoMAEMA} & 181 & 31.50  \\
     & & $\text{VideoMAE-LITE}_{90}$  & 159 & 31.30  \\
     & & $\text{VideoMAE-LITE}_{70}$  & 118 & 30.83  \\
    \bottomrule
    
  \end{tabular}%
  }
  \caption{ Generalization performance of the selector trained on Kinetics-400 and tested on the different datasets and tasks. The results of VideoMAE serve as baseline, which are trained and tested on the same dataset. We use Top-1 accuracy as the metric for classification task and mAP for detection task.}
  \label{tab:zeroshot}
  \vspace{-0.3cm}
\end{table}

\subsection{Qualitative Results}

\begin{figure}
    \centering
    \includegraphics[width=1\columnwidth]{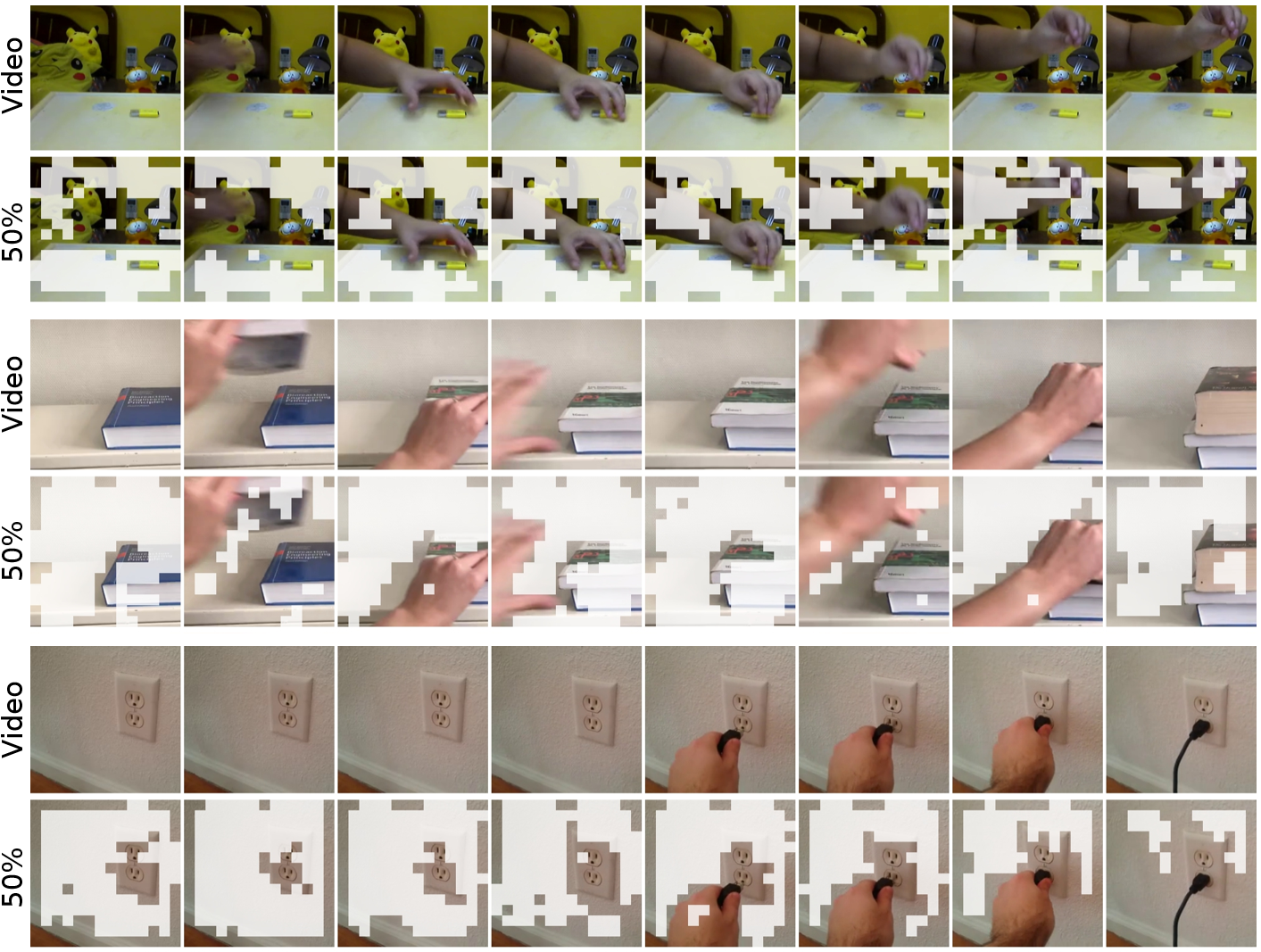}
    \caption{ Visualization of selected tokens at 50\% for different classes, including ``Pretending to pick something up", ``Stacking number of something", and ``Plugging something into something".}     
    \label{fig:sample_results}
    \vspace{-0.5cm}
\end{figure}

Figure~\ref{fig:sample_results} illustrates the token selection of LITE. 
In conjunction with the results presented in Tables~\ref{tab:sota_ssv2} and~\ref{tab:sota_k400}, it is evident that \model effectively selects tokens beneficial to the model and discards those that could mislead classification. 
Additionally, observing the tokens selected by the \oracle reveals that the combination of effective tokens is diverse rather than concentrated. This diversity partly explains why the simple MLP selector is effective in our method; its simplicity helps prevent the \oracle from overfitting while preserving the ability to select a diverse range of tokens.

\subsection{Analysis of \model}
\label{sec:ablation}

\begin{table}
  \centering
  \resizebox{0.92\columnwidth}{!}{%
  \begin{tabular}{lccccc}
    \toprule
    Model & P-Ratio & \gflops & Top-1 & Top-5 \\
    \midrule
    VideoMAE (baseline) & 1 & 181 & 70.50 & 92.35  \\
    Oracle (true label) & 0.5 & 80 & 78.52 & 94.85 \\
    Oracle (predicted label) & 0.5 & 80 & 70.00 & 91.25 \\
    \bottomrule
  \end{tabular}%
  }
  \caption{ Results using different labels to compute the gradient for the oracle. Using the true label and sampling 50\% of tokens produces much better results than even using all tokens. }
  \label{tab:ablation_oracle_ssv2}
  \vspace{-0.2cm}
\end{table}

\begin{table}
  \centering
  \resizebox{0.38\textwidth}{!}{%
  \begin{tabular}{lcl}
    \toprule
    Model & \gflops $\times$ \text{views} & Top-1 \\
    \midrule
     $\text{VideoMAE-LITE}_{90}$  & 159x2x3\textsubscript{\phantom{\textcolor{blue}{$\downarrow$00\%}}} & 70.54 \\
     $\text{VideoMAE-LITE++}_{90}$  & 105x2x3\textsubscript{\textcolor{blue}{$\downarrow$34\%}} & 69.94\textsubscript{\textcolor{blue}{$\downarrow$0.6}}\\
     \midrule
    $\text{VideoMAE-LITE}_{70}$  & 118x2x3\textsubscript{\phantom{\textcolor{blue}{$\downarrow$00\%}}} & 70.44 \\
    $\text{VideoMAE-LITE++}_{70}$  & 84x2x3\textsubscript{\textcolor{blue}{$\downarrow$29\%}} & 69.86\textsubscript{\textcolor{blue}{$\downarrow$0.6}}  \\
    \midrule
    $\text{VideoMAE-LITE}_{50}$  & 80x2x3\textsubscript{\phantom{\textcolor{blue}{$\downarrow$00\%}}} & 69.91  \\
    $\text{VideoMAE-LITE++}_{50}$  & 65x2x3\textsubscript{\textcolor{blue}{$\downarrow$19\%}} & 69.44\textsubscript{\textcolor{blue}{$\downarrow$0.5}}  \\
    \midrule
    $\text{VideoMAE-LITE}_{30}$  & 46x2x3\textsubscript{\phantom{\textcolor{blue}{-00\%}}} & 68.33  \\
    $\text{VideoMAE-LITE++}_{30}$  & 40x2x3\textsubscript{\textcolor{blue}{$\downarrow$13\%}} & 67.74\textsubscript{\textcolor{blue}{$\downarrow$0.6}} \\
    \bottomrule
  \end{tabular}%
  }
  \caption{ Result comparison between LITE and LITE++. In blue the reduced percentage of \gflops and accuracy.}
  \label{tab:lite_adaptive}
  \vspace{-0.3cm}
\end{table}

\myparagraph{Oracle creation.} Gradient scores in the oracle can be calculated in two ways: using the true label or the highest predicted class. Table~\ref{tab:ablation_oracle_ssv2} shows their performance with 50\% of tokens. We observe that using the true label provides a very large improvement of 8.5\%. 

\myparagraph{Adaptive budget.} We experiment with the adaptive budget and show results in Table~\ref{tab:lite_adaptive}. We observe that this is a promising direction as the computation can be reduced by up to a third, while sacrificing $<1\%$ in accuracy. %
We term this model LITE++, enabling further computational savings (additional results in Supplemental Material).

\myparagraph{Experiments across backbones.} 
We explore the effect of using a different backbone, the TimeSformer~\cite{Bertasius2021IsSA} and show results in Table~\ref{tab:ours_res_ssv2}. While for high P-Ratio values \model can still preserve accuracy while reducing a large percentage of GFLOPs, the effect is less striking than using VideoMAE. This points to the fact that a less accurate backbone leads to a noisier oracle, affecting the overall performance of LITE. 

\begin{table}
  \centering
  \resizebox{0.42\textwidth}{!}{%
  \begin{tabular}{lccccc}
    \toprule
    Model & P-Ratio & \gflops & Top-1 & Top-5 \\
    \midrule
    TimeSformer-B & 1 & 180 & 56.23 & 83.63  \\
    TimeSformer-\model & 0.9 & 158 & 56.00 & 83.37 \\
    TimeSformer-\model & 0.7 & 117 & 53.38 & 81.62 \\
    TimeSformer-\model & 0.5 & 79 & 48.28 & 77.62 \\
    TimeSformer-\model & 0.3 & 45 & 36.97 & 66.52 \\
    \bottomrule
  \end{tabular}%
  }
  \caption{LITE's results on a different backbone on \ssv dataset.}
  \label{tab:ours_res_ssv2}
  \vspace{-0.3cm}
\end{table}

\section{Conclusion}
\label{sec:conclusion}

We took a close look at the nature of video tokens to distill five interesting insights about what makes a token valuable for video understanding, how this value is distributed and how the number of tokens needed correlates to the difficulty of the task. We leverage these principles to design a token selector, which we call LITE, that outperforms the existing state-of-the-art in the trade-off between accuracy and computation. We hope both the principles as well as \model can cast a new light into research on efficient Transformer-based video analysis models.

{
    \small
    \bibliographystyle{ieeenat_fullname}
    \bibliography{main}
}

\end{document}